\documentclass[lettersize,journal]{IEEEtran}
\usepackage{amsmath,amsfonts}
\usepackage{autobreak}
\usepackage{algorithmic}
\usepackage{algorithm}
\usepackage{array}
\usepackage[caption=false,font=normalsize,labelfont=sf,textfont=sf]{subfig}
\usepackage{textcomp}
\usepackage{stfloats}
\usepackage{url}
\usepackage{verbatim}
\usepackage{graphicx}
\usepackage{cite}
\usepackage{hyperref}
\usepackage{graphics}
\usepackage{epsfig}
\usepackage{gensymb}
\usepackage{algorithm} 
\usepackage{algorithmic} 
\usepackage{multirow} 
\usepackage{utfsym}
\usepackage{tabularx}
\usepackage{booktabs}
\begin{document}

\title{TRLO: An Efficient LiDAR Odometry with 3D Dynamic Object Tracking and Removal}

\author{Yanpeng Jia, Ting Wang$^{*}$, Xieyuanli Chen, Shiliang Shao
\thanks{*This work was supported by National Natural Science Foundation of China (Grant No. 62203091 and U20A20201), the China postdoctoral Science Foundation (Grant No. GZB20230804), Autonomous Project of State Key Laboratory of Robotics (Grant No. 2024-Z09). (\textit{Corresponding author: Ting Wang})}
\thanks{Y. Jia, T. Wang and S. Shao are with the State Key Laboratory of Robotics at Shenyang Institute of Automation, Chinese Academy of Sciences, Shenyang, China. X. Chen is with the College of Intelligence Science and Technology, National University of Defense Technology, Changsha, China. Y. Jia is also with the University of Chinese Academy of Sciences, Beijing, China. (email: wangting@sia.cn)}}

\markboth{Journal of \LaTeX\ Class Files,~Vol.~14, No.~8, August~2021}%
{Shell \MakeLowercase{\textit{et al.}}: A Sample Article Using IEEEtran.cls for IEEE Journals}

\IEEEpubid{0000--0000/00\$00.00~\copyright~2021 IEEE}

\maketitle

\begin{abstract}
Simultaneous state estimation and mapping is an essential capability for mobile robots working in dynamic urban environment. The majority of existing SLAM solutions heavily rely on a primarily static assumption. However, due to the presence of moving vehicles and pedestrians, this assumption does not always hold, leading to localization accuracy decreased and maps distorted. To address this challenge, we propose TRLO, a dynamic LiDAR odometry that efficiently improves the accuracy of state estimation and generates a cleaner point cloud map. To efficiently detect dynamic objects in the surrounding environment, a deep learning-based method is applied, generating detection bounding boxes. We then design a 3D multi-object tracker based on Unscented Kalman Filter (UKF) and nearest neighbor (NN) strategy to reliably identify and remove dynamic objects. Subsequently, a fast two-stage iterative nearest point solver is employed to solve the state estimation using cleaned static point cloud. Note that a novel hash-based keyframe database management is proposed for fast access to search keyframes. Furthermore, all the detected object bounding boxes are leveraged to impose posture consistency constraint to further refine the final state estimation. Extensive evaluations and ablation studies conducted on the KITTI and UrbanLoco datasets demonstrate that our approach not only achieves more accurate state estimation but also generates cleaner maps, compared with baselines.
\end{abstract}


\section{Introduction} \label{sec1}

\IEEEPARstart{A}{ccurate} state estimation is indispensable for autonomous vehicles to localize themselves and explore in unknown environments without GPS assistance. In contrast to visual SLAM methods \cite{vins,orb-slam}, which may perform poorly under varying illumination or uniform textures, LiDAR-based methods can provide more reliable pose estimation by precisely capturing the geometric details of the environment \cite{survey}. Feature-based approaches \cite{loam,lego-loam,lio-sam} attempt to solve the transformation between two adjacent scans by focusing on the most representative points, such as edges or planes. However, they may inadvertently discard well-structured points which could help improve the quality of downstream registration. Especially, when a scene lacks obvious features, the feature-based method may struggle to extract sufficient feature points, which may lead to pose drift and mapping distortion.

\begin{figure}[tbp]
	\centering
	\includegraphics[width=0.5\textwidth]{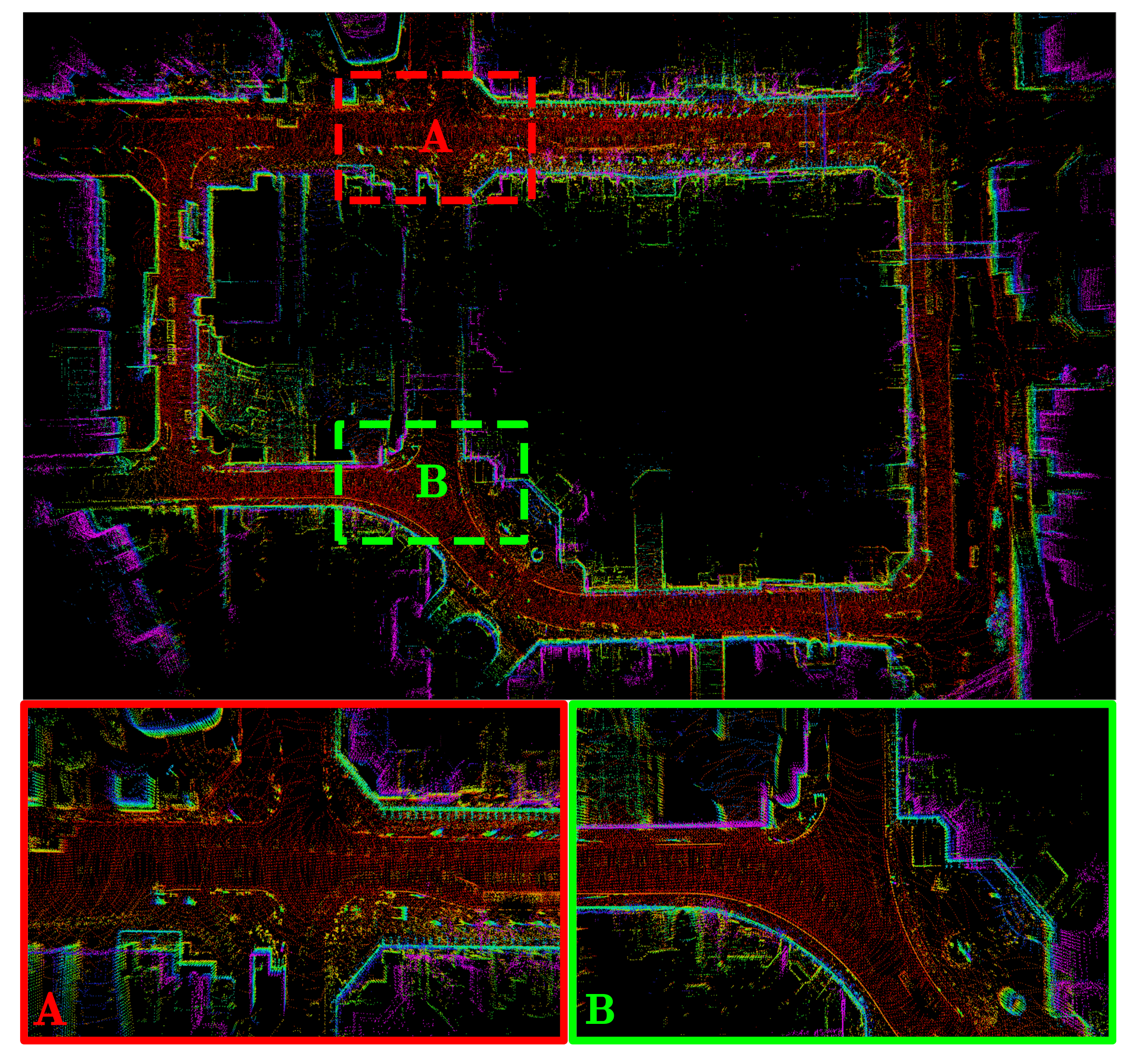}
	\caption{\textbf{Our global clean map with dynamic objects removal on Urbanloco dataset}. A and B are two examples of local map details at crowded traffic intersections, presenting that the superiority of our method on mapping.}
	\label{figure1}
\end{figure}

Nowadays, an alternative access directly utilizing the entire point cloud to estimate the transformation between LiDAR scans, typically by iterative closest point (ICP) \cite{icp}, is popular. However, when dealing with dense scans, ICP technique suffers from prohibitively computational consume, which may affect real-time performance. To address this problem, various ICP variants \cite{v-icp,g-icp} are proposed to improve the computational efficiency. Some derived LiDAR odometry (LO) works \cite{dlo,ct-icp,kiss-icp} have achieved promising performance based on the improved ICP algorithms.

\IEEEpubidadjcol

The majority of existing LO solutions \cite{loam,lego-loam,dlo,ct-icp,kiss-icp} heavily rely on the basic assumption of a static environment. Although some of them \cite{ct-icp,kiss-icp} mitigate the impact of dynamic objects through some algorithm designs, they do not directly focus on moving objects, leading to the localization accuracy decline or potential failure. In addition, moving objects often affect the quality of mapping and hinder following navigation tasks. Therefore, effectively detecting and tracking dynamic objects has become a key for robots' robust localization and mapping in real-world.

Although there exist some impressive works \cite{rf-lio,dor-lins,suma++,dl-slot,lio-segmot} introducing innovative methods to tackle SLAM challenges in dynamic environments, they still have some limitations. RF-LIO \cite{rf-lio}, DOR-LINS \cite{dor-lins} and LIMOT \cite{limot} relies on IMU to provide pose priors, which may be susceptible to noise. Many works \cite{suma++,doa} incorporate strict semantic segmentation into LO system, whose performances are limited with sparse-channel LiDAR. \cite{dl-slot} and \cite{lio-segmot} both focus on improving the accuracy of object tracking rather than reducing the impact of dynamic objects on the accuracy of the LiDAR odometry.

In this work, a novel and efficient dynamic LO, TRLO, is proposed. To efficiently detect and track dynamic objects, a pillars-based object detection network is implemented as a TensorRT module, and a multi-object tracking (MOT) method based on Unscented Kalman Filter (UKF) \cite{ukf} and nearest neighbor (NN) strategy is proposed. Subsequently, the static point cloud is fed into Fast G-ICP \cite{fast-icp} module for two-stage scan-matching to achieve accurate localization and mapping. Furthermore, we apply the bounding boxes accumulated in a sliding window to impose consistency constraints, further refining the pose of robot. Fig.~\ref{figure1} shows an example of point cloud map generated with our proposed method. The contributions of our work are summarized as follows:

\begin{itemize}
	\item We propose TRLO, a LiDAR odometry that can accurately detect and track dynamic objects for high-precision state estimation and clean mapping, whose key is the proposed reliable 3D multi-object tracker based on UKF and NN.
	\item We propose a hash-based keyframe database management to accelerate scan-to-map registration through direct access to the keyframe index.
	\item We reuse the postures of bounding boxes accumulated in a short-term sliding window to impose consistency constraint on state estimation, assuming that all the objects stand on the same ground flat in the local scene.
	\item Extensive experiments conducted on challenging sequences of the KITTI and UrbanLoco datasets demonstrate that our method outperforms baselines on localization and mapping. The source code and a demo video of our approach are available at: \url{https://yaepiii.github.io/TRLO/}.
\end{itemize}

\section{Related work} \label{sec2}

LiDAR odometry and SLAM are classic topics in robotics with a quantity of researches. In this section, we briefly review traditional LiDAR odometry estimation methods, and focus on SLAM solutions in dynamic scenarios.

\subsection{Traditional LiDAR Odometry}	\label{sec2a}

As the most typical feature-based method, LOAM \cite{loam} introduces an innovative approach by separating the estimation problem into two distinct algorithms that one operates at high frequency with low accuracy, while the other operates at low frequency with high accuracy. The two results are fused together to produce a final estimate at both high frequency and high accuracy. LeGO-LOAM \cite{lego-loam} achieves lightweight LiDAR odometry by using ground point segmentation and a two-stage Levenberg-Marquardt algorithm, and improves accuracy using culling mechanism for dynamic small objects. MULLS \cite{mulls} concentrates on employing principal component analysis (PCA) to extract more abundant and stable features, which proposes a LiDAR-based SLAM framework independent of scan lines, thereby enhancing generalization. Li et al. \cite{ael} propose a novel angle-based feature extraction method based on the equiangular distribution property of the point cloud, and construct a fixed-lag smoothing to jointly optimize the poses of multiple keyframes.

With the rapid iterative advancements in processing units \cite{task}, some recent works are exploring direct utilization of ICP or its variants for real-time state estimation and mapping. Direct LiDAR Odometer (DLO) \cite{dlo}, introducing NanoGICP, a lightweight iterative nearest point solver, facilitates precise registration of dense original point clouds to determine the robot's pose. It realizes reasonable keyframe management and mapping through an adaptive method. CT-ICP \cite{ct-icp} introduces the combined continuity in the scan matching and discontinuity between scans to enhance the accuracy and robustness of point cloud registration. KISS-ICP \cite{kiss-icp} is the latest method of direct registration, which focuses on simplifying the implementation of ICP. It achieves a balance between accuracy and speed by employing adaptive threshold data association.

\subsection{LiDAR-based SLAM in Dynamic Scenarios}	\label{sec2b}

Traditional methods operate under the predominant assumption of a static environment. However, in dynamic scenarios such as dynamic urban environments, the decline of odometry accuracy and mapping distortion may happen. Several SLAM schemes endeavor to identify and filter dynamic objects within the environment to safeguard location and mapping accuracy.

Early efforts \cite{dynamicslam,cd-slam} use visual techniques to identify and remove dynamic objects. In order to achieve outdoor large-scale dynamic LiDAR-based SLAM, certain combines 3D LiDAR sensors with other sensors. Based on the LIO-SAM \cite{lio-sam} framework, RF-LIO \cite{rf-lio} uses the initial value provided by IMU and multi-resolution depth map to detect and remove dynamic points in submap, and achieves precise registration based on scan-to-static submap. DOR-LINS\cite{dor-lins} introduce Ground Pseudo Occupancy to identify and remove dynamic objects. However, the introduction of other sensors will lead to expensive cost and extra complexity. In contrast, our approach uses LiDAR as the only sensor, mitigating expenses while achieving comparable or even better results.

Motivated by the advances of deep learning and Convolutional Neural Networks (CNNs) for scene understanding \cite{rgb-di,rss}, some methods extract semantic information from LiDAR data for dynamic removal. SuMa++ \cite{suma++} uses RangeNet++ \cite{rangenet++} to conduct spherical projection and extract semantic information for dynamic removal through point-wise probabilistic method. Pfreundschuh \textit{et al}. \cite{doa} uses deep learning techniques for real-time 3D dynamic object detection. They extract features from static point clouds to estimate its own motion. Despite effective, these methods impose constraints on the generalization of LiDAR odometry, as semantic extraction networks are tailored specifically for 64-channel or denser LiDAR. Consequently, when employing 32-channel LiDAR, the system becomes inoperative. Our method still yields surprising results on the datasets using 32-channel LiDAR.

Reference \cite{dl-slot} and \cite{lio-segmot} are most similar to our work. They both integrate the objects' pose into the factor graph estimation framework for optimization, so as to achieve effective tracking of the dynamic objects. However, \cite{dl-slot} removes all detected objects before the odometry estimation, neglecting the valuable static information. Due to asynchronous state estimation, \cite{lio-segmot} does not operate in real-time. Although S$^2$MAT \cite{s2mat} focuses on dynamic object tracking and removal, it does not address how to mitigate the impact of dynamic objects on the LiDAR odometry to improve localization accuracy. Additionally, some methods \cite{removert,erasor} focus on map construction in dynamic scenarios and therefore generally do not emphasize real-time.

\begin{figure*}[tbp]
	\centering
	\includegraphics[width=\textwidth,height=7cm]{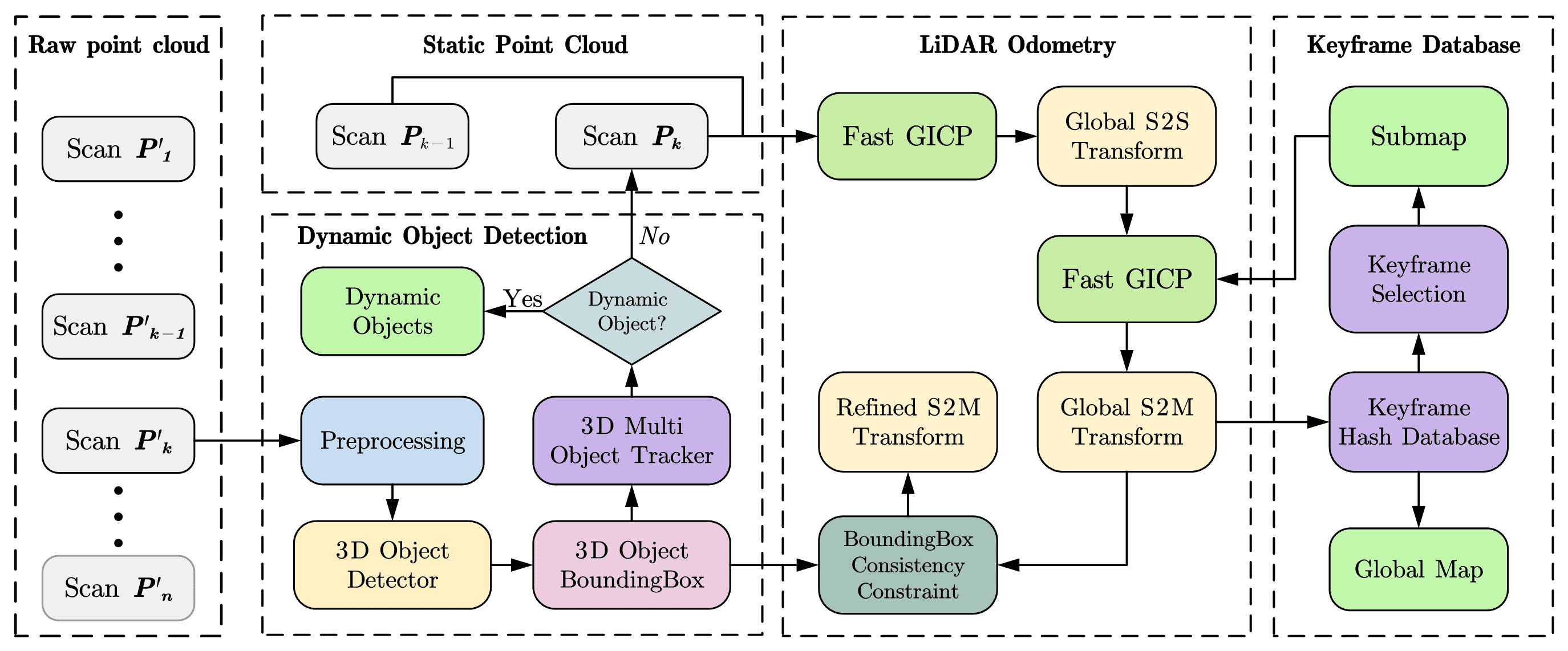}
	\caption{\textbf{Our system architecture.} The pillars-based 3D object detector first is used for preprocessing of the raw point cloud to detect dynamic and semi-static objects, resulting in the generation of a 3D object bounding boxes. Subsequently, 3D multi-object tracker is applied to identify and remove dynamic objects. The adjacent static scans $P_k$ are input to calculate the scan-to-scan(S2S) transformation. The initial value is propagated to the world frame and used for the secondary Fast GICP of scan-to-map(S2M). $P_k$ is scan-matched with the submap $S_k$ composed of selective keyframes. Finally, the S2M transformation is further optimized with the posture consistency constraint imposed by the detected bounding boxes to obtain a refined global robot's pose, which is checked against multiple metrics to determine if it should be stored in keyframe hash dataset.}
	\label{figure2}
\end{figure*}

\section{Method} \label{sec3}

In this paper, our system architecture is shown in Fig.~\ref{figure2}. Our work attempts to address the following problem: Given adjacent raw point cloud scans $P'_k$ and $P'_{k-1}$ at time $k$, remove dynamic objects to obtain static point cloud $P_{k}$ and $P_{k-1}$ as the input of LiDAR odometry, and then estimate the robot's global pose $ \chi^{\mathcal{W}}_k \in SE(3)$ and build a global map $M_k$ in the coordinate system of world $\mathcal{W}$, where  $\chi_k$ consists of rotation part $R_k \in SO(3)$ and translation part $t_k \in \mathbb{R}^3$.

\begin{figure}[tbp]
	\centering
	\includegraphics[width=0.485\textwidth]{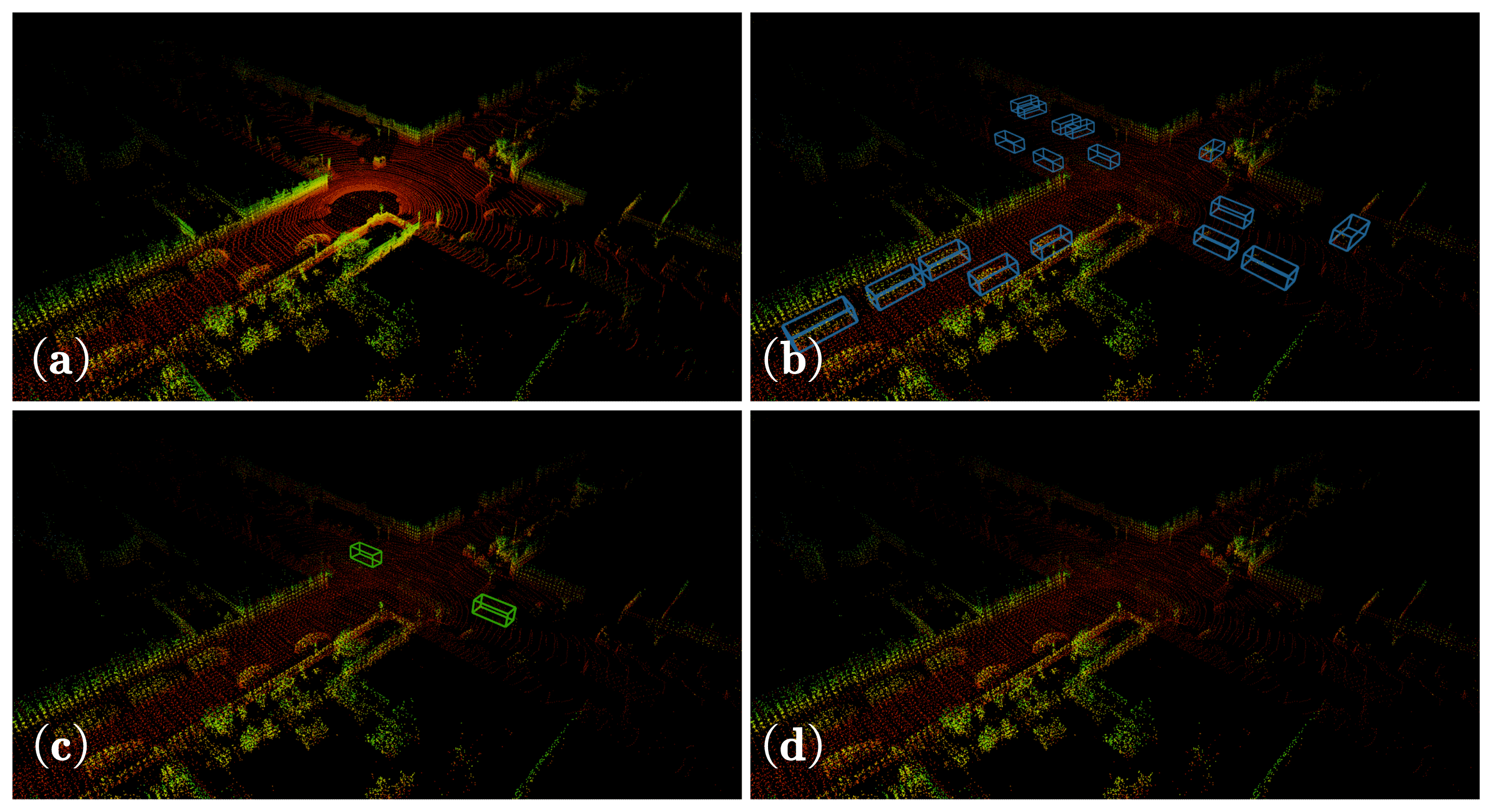}
	\caption{\textbf{Effect of the proposed detection and filtration of dynamics.} (a) Raw point cloud. (b) 3D bounding boxes of dynamic and semi-static objects detected by our 3D object detector (blue boxes). (c) 3D bounding boxes of dynamic objects identified by our 3D multi-object tracker (green boxes).  (d) Static point cloud after removing the dynamic objects, which will be input for LiDAR Odometry.}
	\label{figure3}
\end{figure}

\subsection{3D Object Detector} \label{sec3a}
When our system receives the raw point cloud $P_k'$ collected by $360 \degree$ LiDAR (Fig.~\ref{figure3}(a)), in order to minimize the noise and the loss of raw data to ensure the accuracy of object detection, only 0.5m box filter around the robot and 0.25m voxel filter are used to reduce false detection and remove noise in the preprocessing stage. The processed point cloud is then fed into the 3D object detector.

In this module, we use PointPillars \cite{pointpillars} as the backbone for feature map extraction, and then use Center-based \cite{cpp} to generate 3D bounding boxes (Fig.~\ref{figure3}(b)) for dynamic and semi-static objects in which semi-static objects are defined as those potentially moving at the current time. Notably, although we chose Center-base \cite{cpp} that guarantees a trade-off between precision and speed, our pipeline is not limited to a specific object detection method. Let $B^{\mathcal{L}}_{k,m}$ be a bounding box in LiDAR coordinate system $\mathcal{L}$ at $k$ scan all bounding boxes $B^{\mathcal{L}}_{k}=\{B^{\mathcal{L}}_{k,m} | m=1,...,M\}$. To mitigate the impact of false detection on 3D multi-object tracking, we selectively detect cars and cyclists while raising the threshold to 0.75 following \cite{sort}. Additionally, we optimize the entire efficiency of the system by rewriting 3D object detector as a TensorRT version and integrating it into the system front-end, resulting in more than double the processing speed.

\subsection{3D Multi-Object Tracker} \label{sec3b}
Traditional SLAM systems rely on geometric information to locate and associate observations to the map. However, they fail to account for dynamic scenes changes caused by moving objects, which may lead to incorrect registration and map distortion. Dynamic SLAM algorithms are usually equipped with certain outlier rejection. However, if we simply discard dynamic and semi-static objects, we risk losing their valuable contribution to odometry estimation.

We present an approach based on UKF and NN technique to track the detected bounding boxes $B^{\mathcal{L}}_{k}$ from \ref{sec3a}. Different from EKF, the UKF addresses the nonlinear state estimation by using a deterministic sampling approach instead of linear approximation \cite{ukf}, more suitable for dynamic object tracking. The deductions of object tracking process are as follows:

1) Estimation Model: We approximate the inter-frame displacement of each object with a constant velocity model that is independent of other objects and agent. For simplicity, we omit the subscript $m$. Given an object $x^{\mathcal{O}}_k$ at $k$th scan, its state is defined as:

\begin{equation}
	\label{eq1}
	x^{\mathcal{O}}_k = [x,y,z,\theta,v,l,w,h]
\end{equation}

where, $x, y, z$ are the center coordinates of the target in the robot coordinate system $\mathcal{B}$. Note that we assume $\mathcal{L}$ and $\mathcal{B}$ coincide. $\theta$ is the yaw Angle between the object coordinate system $\mathcal{O}$ and $\mathcal{B}$, $v$ is the speed that needs to be estimated, $l, w, h$ are the length, width and height of the bounding box, respectively. Then we estimate the state of the each object through UKF \cite{ukf}:

\begin{small}
	\begin{flalign}
		&\ \ \ \text{predict}\left\{ \begin{array}{l}
			\hat{x}_{k+1|k}^{\mathcal{O}}=\sum_{i=0}^{2n}{W_{i}^{m}\mathcal{X}_{k+1|k}^{\mathcal{O}}}\\
			\\
			\hat{P}_{k+1|k}^{\mathcal{O}}=P_{xx,k+1|k}^{\mathcal{O}}\\
		\end{array} \right. & \label{eq2}
	\end{flalign}
	\begin{flalign}
		&\ \ \ \text{update}\left\{ \begin{array}{l}
			\hat{y}_{k+1|k}^{\mathcal{O}}=\sum_{i=0}^{2n}{W_{i}^{m}\mathcal{Y} _{k+1|k}^{\mathcal{O}}}\\
			\\
			K_{k+1}=P_{xy,k+1|k}^{\mathcal{O}}\left( P_{yy,k+1|k}^{\mathcal{O}} \right) ^{-1}\\
			\\
			\hat{x}_{k+1|k+1}^{\mathcal{O}}=\hat{x}_{k+1|k}^{\mathcal{O}}+K_{k+1}\left( y_{k+1}-\hat{y}_{k+1|k}^{\mathcal{O}} \right)\\
			\\
			\hat{P}_{k+1|k+1}^{\mathcal{O}}=\hat{P}_{k+1|k}^{\mathcal{O}}+K_{k+1}P_{yy,k+1|k}^{\mathcal{O}}K_{k+1}^{T}\\
		\end{array} \right. & \label{eq3}
	\end{flalign}
\end{small}
where, $\mathcal{X}^{\mathcal{O}}_{k+1|k}$ and $\mathcal{Y}^{\mathcal{O}}_{k+1|k}$ are the values obtained after proportional modified symmetric sampling and nonlinear transfer, $W_{i}^{m}$ and $W_{i}^{c}$ are the corresponding weights, $K_{k+1}$ is the Kalman gain, $y_{k}$ is observation at $k$ scan, $\hat{x}_{k+1|k+1}^{\mathcal{O}}$ and $\hat{P}_{k+1|k+1}^{\mathcal{O}}$ are the estimated state and covariance, respectively, $P_{xx,k+1|k}^{\mathcal{O}}$, $P_{xy,k+1|k}^{\mathcal{O}}$, and $P_{yy,k+1|k}^{\mathcal{O}}$ are the covariance matrices of $\mathcal{X}^{\mathcal{O}}_{k+1|k}$, $\mathcal{Y}^{\mathcal{O}}_{k+1|k}$. More details are referred to \cite{ukf}. According to the estimated speed of objects, we define the tracked objects as dynamic if their estimated speed is over 1m/s, otherwise the tracked objects are regarded as semi-static objects. Our method effectively distinguishes dynamic objects (Fig.~\ref{figure3}(c)) or semi-static objects based on the estimated velocity.

2) Data Association: After estimating the new state of each target using \eqref{eq2} and \eqref{eq3}, we perform the data association step. Instead of employing the typical IOU-based criteria, we simply use the nearest neighbor technique to assign targets to the detections in the current frame because we find that this simple approach also achieves high accuracy in the application. When a detection is associated to a target, the detected bounding box is used to update the target state. Additionally, to mitigate the impact of misdetection or occlusion, if no target is associated with a detection, we only perform the prediction step without update.

3) Creation and Deletion of Tracking Targets: When objects enter the LiDAR field of view, the tracker is initialized using the geometry of the bounding box with the velocity set to zero. Since the velocity is unobserved, the covariance of the velocity component is initialized with large values, reflecting this uncertainty. When an object leaves the LiDAR field of view, rather than deleted immediately, it will be kept for a $Age_{max}$ lifetime during which only prediction step is performed. This approach effectively avoids the problem of deleting the true positive trajectory that still exists in the scene but cannot be matched due to misdetection or occlusion. As shown in Fig.~\ref{figure3}(d), we obtain a cleaner static point cloud after dynamic removal, while keeping the semi-static objects that may provide valuable information for LiDAR odometry.

\subsection{LiDAR Odometry} \label{sec3c}

LiDAR Odometry can be viewed as the process of recovering $SE(3)$ transformation by comparing the processed adjacent static point cloud with the static point clouds in memory. The process is usually performed in two stages, first to provide a best instantaneous guess, followed by refinement to make it more globally consistent.

1) Scan-to-Scan: In the first stage, the scan-to-scan aims to compute the relative transform $\hat{\chi}^{\mathcal{L}}_k$ between a source point cloud $P^s_k$ and target point cloud $P^t_k$ (where $P^t_k = P^s_{k-1}$) captured in $\mathcal{L}$ where

\begin{equation}
	\hat{\chi}^{\mathcal{L}}_k = \underset{\chi^{\mathcal{L}}_k}{\text{arg}\min}\ \varepsilon\left( \chi^{\mathcal{L}}_k P_{k}^{s},\ P_{k}^{t} \right)                  \label{eq4}
\end{equation}
the residual error $\varepsilon$ from Fast GICP is defined as

\begin{equation}
	\varepsilon\left( \chi^{\mathcal{L}}_k P_{k}^{s},\ P_{k}^{t} \right) =  \sum_i^N{d_{i}^{T}\left( C_{k,i}^{t}\ +\ \chi^{\mathcal{L}}_k C_{k,i}^{s} \chi^{\mathcal{L}^T}_k \right) ^{-1}d_i}
	\label{eq5}
\end{equation}

such that the overall objective for this stage is
\begin{equation}
	\hat{\chi}^{\mathcal{L}}_k =  \sum_i^N{d_{i}^{T}\left( C_{k,i}^{t}\ +\ \chi^{{\mathcal{L}}^T}_k C_{k,i}^{s} \chi^{\mathcal{L}}_k \right) ^{-1}d_i}
	\label{eq6}
\end{equation}
where, N is the number of corresponding points between source $P_k^s$ and target $P_k^t$, $d_i = p_i^t - \chi^{\mathcal{L}}_k p_i^s$, $p_i^s \in P_k^s$, $p_i^t \in P_k^t$, and $C^s_{k,i}$ and $C^t_{k,i}$ are the estimated covariance matrices corresponding to each point $i$ of the source and target cloud, respectively. If an external sensor, such as an IMU, is available, our system can provide an initial guess for Fast GICP through IMU pre-integration. However, it is important to note that our system is not reliant to IMU. In case where IMU is unavailable, we can still achieve a robust state estimation by setting the initial guess to the identity matrix $I$.

Finally, we accumulate the result $\hat{\chi}^{\mathcal{L}}_k$ to the previous global scan-to-map transform $\hat{\chi}^{\mathcal{W}}_{k-1}$, i.e. $\hat{\chi}^{\mathcal{W}}_k = \hat{\chi}^{\mathcal{W}}_{k-1}\hat{\chi}^{\mathcal{L}}_k$.

2) Scan-to-Map: In this stage, we use $\hat{\chi}^{\mathcal{W}}_k$ as the initial guess of scan-to-map and follow a similar procedure as scan-to-scan. However, the objective here is to obtain a globally consistent robot global state $\hat{\chi}^{\mathcal{W}}_k$ by matching the current static point cloud $P^s_k$ with the local submap $S_k$ such that

\begin{equation}
	\hat{\chi}^{\mathcal{W}}_k = \underset{\chi^{\mathcal{W}}_k}{\text{arg}\min}\ \varepsilon\left( \chi^{\mathcal{W}}_k P_{k}^{s},\ S_k \right)
	\label{eq8}
\end{equation}

With the same residual $\varepsilon$ defined in \eqref{eq5}, then the overall objective function of scan-to-map stage 

\begin{equation}
	\hat{\chi}^{\mathcal{W}}_k =  \sum_j^M{d_{j}^{T}\left( C_{k,j}^{S}\ +\ \chi^{{\mathcal{W}}^T}_k C_{k,j}^{s} \chi^{\mathcal{W}}_k \right) ^{-1}d_j}
	\label{eq9}
\end{equation}
where, $M$ is the number of corresponding points between source $P_k^s$ and submap $S_k$, and $C^S_{k,j}$ is the corresponding scan-stitched covariance matrix for each point $j$ of the submap as defined later in \ref{sec3d}.

3) Bounding Box Consistency Constraint: We use 3D bounding boxes generated in \ref{sec3a} to impose posture consistency constraint on $[t_z,roll,pitch]$ of scan-to-map result. Specifically, all the detected objects are presumed to stand on the same flat ground. We therefore project the center points of the bounding boxes vertically, yielding the ground points of the objects. Then we utilize all these ground points for ground fitting. If the number of interior points exceeds a specified threshold, we leverage them to compute the ground normal, which serves as a constraint for $[roll,pitch]$. It's worth noting that in certain scans with few bounding boxes, this process may fail. To avoid this problem, we implement a sliding window to accumulate the bounding boxes of several consecutive scans. All bounding boxes within this sliding window are utilized for ground fitting.

Additionally, if the mean change $z$-axis of the bounding boxes between adjacent scans is less than 0.1m, we infer that the robot is traversing relatively flat terrain. In this case, we presume the ground to not change between two consecutive scans by considering the $z$-axis displacement in the local window close to zero, thereby imposing constraint on $t_z$.

\subsection{Hash-Based Keyframe Management} \label{sec3d}

Inspired by the visual SLAM approach \cite{orb-slam}, we use keyframes to manage map. However, different from the common keyframe management, we use hash table to access keyframes efficiently. Referring to the idea of adaptive threshold of DLO \cite{dlo}, the threshold on adding keyframe is adaptively adjusted according to the \textit{spaciousness} of the current environment. Please refer to \cite{dlo} for more details. When the robot's moving distance or rotation angle is greater than the threshold, we associate the robot's current state $\chi_k$ with the static point cloud $P_k$ in key-value pairs, and then store it into the hash-based keyframe database.

When submap needs to be built, we implement a keyframe selection strategy based on nearest neighbor and convex hull, concave hull. Let $K_k$ be the set of all keyframe point clouds, and the submap $S_k$ is defined as the series of $K$ nearest neighbor keyframe scans $Q_k$, $L$ nearest neighbor convex hull scans $H_k$ and $J$ nearest neighbor concave hull $G_k$ such that $S_k = Q_k \oplus H_k \oplus G_k$, where $\oplus$ represents the intersection. This combination reduces overall trajectory drift by maximizing scan-to-map overlap and provides the system with multiple scales of environmental features to align.

\section{Experiments} \label{sec4}

To validate the effectiveness of our method, we rigorously conduct extensive experiments on the  KITTI \cite{kitti} and UrbanLoco \cite{ulhk} datasets, including odometry benchmark and mapping results, as well as running time analysis and ablation study of each component. All experiments are executed on a laptop with a 20-core Intel i9 2.50GHz CPU and a RTX 3060 GPU.

\begin{table*}[ht]
	\caption{Odometry Benchmark Results Using RMSE of RPE (m) and APE (m) on KITTI and UrbanLoco datasets}
	\centering
	\scalebox{1.07}{
		\begin{tabular}{|c|c|c|c|c|c|c|c|c|c|c|c|c|}
			\hline
			\rule{-1pt}{10pt}
			\multirow{2}{*}{\textbf{Dataset}} & \multicolumn{6}{c|}{\textbf{KITTI}} & \multicolumn{6}{c|}{\textbf{UrbanLoco}}\\
			
			\cline{2-13}
			\rule{-1pt}{10pt}
			
			& \multicolumn{2}{c|}{KITTI00} & \multicolumn{2}{c|}{KITTI05} & \multicolumn{2}{c|}{KITTI07} & \multicolumn{2}{c|}{UrbanLoco01} & \multicolumn{2}{c|}{UrbanLoco03} & \multicolumn{2}{c|}{UrbanLoco05} \\
			\hline
			\rule{-1pt}{8pt}
			
			Dynamic Level & \multicolumn{2}{c|}{low} & \multicolumn{2}{c|}{low} & \multicolumn{2}{c|}{medium} & \multicolumn{2}{c|}{high} & \multicolumn{2}{c|}{medium} & \multicolumn{2}{c|}{high} \\
			\hline
			\rule{-1pt}{8pt}
			
			Method & RPE & APE & RPE & APE & RPE & APE & RPE & APE & RPE & APE & RPE & APE \\
			\hline
			\rule{-1pt}{10pt}
			
			A-LOAM & 1.26 & 8.07 & 1.17 & 4.19 & 1.10 & 0.70 & 5.65 & 1.83 & 3.61 & 2.15 & 3.84 & 15.19 \\
			\hline
			\rule{-1pt}{10pt}
			
			LeGO-LOAM & 1.96 & 6.78 & 1.31 & 3.04 & 1.63 & 0.85 & 5.80 & 2.90 & 3.04 & 1.87 & 3.53 & 21.45 \\
			\hline
			\rule{-1pt}{10pt}
			
			DLO & 1.19 & 6.67 & 1.16 & 3.34 & 1.11 & 1.54 & 5.66 & 2.03 & 3.60 & 1.83 & 3.81 & 1.72 \\
			\hline
			\rule{-1pt}{10pt}
			
			CT-ICP & 1.20 & 5.93 & 1.16 & \underline{\textbf{2.16}} & 1.11 & 0.87 & 5.66 & 1.81 & 3.60 & 2.06 & 3.81 & 1.96 \\
			\hline
			\rule{-1pt}{10pt}
			
			KISS-ICP & 1.22 & 8.55 & 1.16 & 3.32 & 1.11 & 0.67 & 5.67 & 4.16 & 3.61 & 3.09 & 3.82 & 1.96 \\
			\hline
			\rule{-1pt}{10pt}
			
			LIO-SEGMOT & 1.32 & 6.67 & 1.24 & 3.13 & 1.32 & 0.84 & 5.32 & 1.69 & 1.94 & 1.81 & 3.47 & 1.59 \\
			\hline
			\rule{-1pt}{10pt}
			
			LIMOT & 1.24 & 6.46 & 1.22 & 3.55 & 1.24 & 0.79 & 5.50 & 1.65 & 1.88 & \textbf{1.77} & 3.49 & 1.34 \\
			\hline
			\rule{-1pt}{10pt}
			
			RF-A-LOAM & 1.31 & 8.42 & 1.20 & 3.55 & \underline{\textbf{0.99}} & \underline{\textbf{0.62}} & 5.45 & 1.74 & 3.61 & 2.11 & 3.85 & 1.47 \\
			\hline
			\rule{-1pt}{10pt}
			
			Ours & \underline{\textbf{1.15}} & \underline{\textbf{3.29}} & \underline{\textbf{1.15}} & 2.36 & 1.10 & 0.92 & \underline{\textbf{5.16}} & \underline{\textbf{1.57}} & \underline{\textbf{1.58}} & \underline{1.80} & \underline{\textbf{3.42}} & \underline{\textbf{1.23}} \\
			\hline
			\multicolumn{13}{l}{\footnotesize The \textbf{bold} font denotes the best results. The \underline{underscore} denotes the best results among LiDAR-only methods.}
		\end{tabular}
	}
	\label{table1}
\end{table*}

\subsection{Odometry Benchmark} \label{sec4a}
1) Experimental Setup: We conduct a benchmark evaluation to compare our method with several traditional LiDAR odometry methods (A-LOAM \cite{loam}, LeGO-LOAM \cite{lego-loam}, DLO \cite{dlo}, CT-ICP \cite{ct-icp} and KISS-ICP \cite{kiss-icp}) and several dynamic LiDAR-inertial odometry methods (LIO-SEGMOT \cite{lio-segmot} and LIMOT \cite{limot}). For fair comparison, loop closure detection module is disabled in LeGO-LOAM \cite{lego-loam}, LIO-SEGMOT \cite{lio-segmot} and LIMOT \cite{limot}. Since SuMa++ \cite{suma++} filters dynamic objects relying on semantic information extraction, it can not run on the dataset collected by the 32-channel LiDAR. We re-implement dynamic LO, RF-A-LOAM integrating range-based dynamic removal. All experiments are conducted on the KITTI \cite{kitti } and UrbanLoco \cite{ulhk} datasets, using Root Mean Square Error (RMSE) of the relative pose error (RPE) and the absolute pose error (APE) as metrics.

2) Experimental Analysis: The results are shown in Table.~\ref{table1}. Compared with traditional LO systems, our approach generally achieves encouraging performance through the proposed dynamic removal and bounding box consistency constraint strategies, especially for the highly dynamic sequences. Additionally, the proposed method is superior to LIO-SEGMOT \cite{lio-segmot} and LIMOT \cite{limot}, apart from that the APE of Urbanloco03 sequence is lower than that of LIMOT \cite{limot}. Notably, our approach can obtain competitive results using only LiDAR scans as input, which demonstrates the effectiveness of dynamic removal and robust registration mechanism.

\begin{figure*}[htbp]
	\centering
	\includegraphics[width=\textwidth]{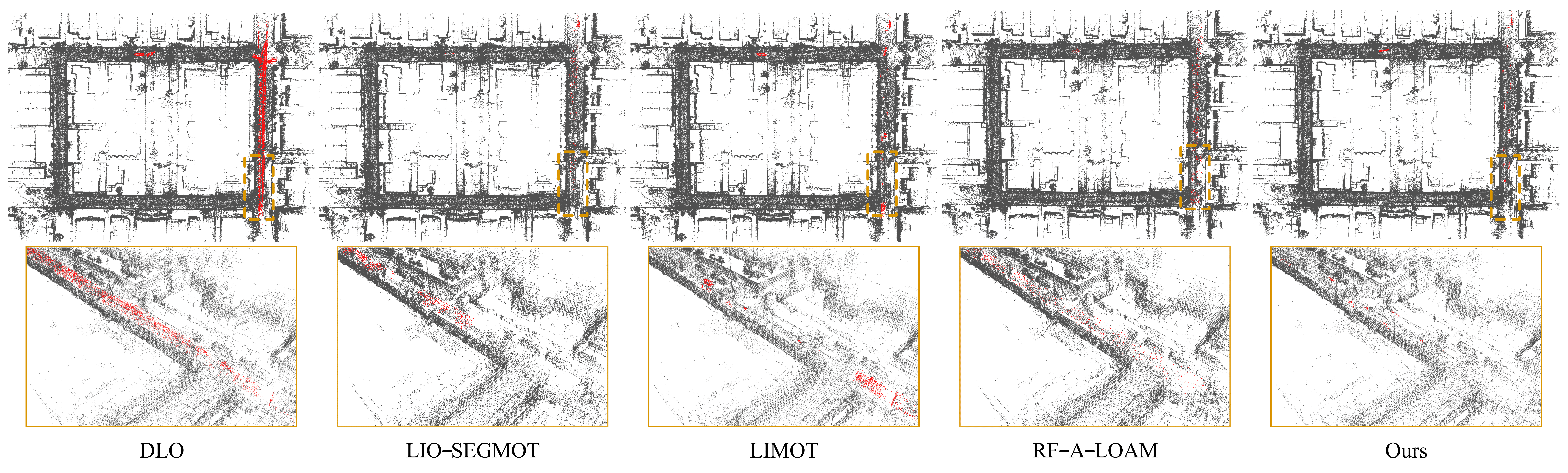}
	\caption{\textbf{Mapping results.} (a) Global map generated by traditional method. (b) Local details of the global map generated by traditional method, where red points represent \textit{ghosttail} caused by moving objects. (c) Global map generated by our method. (d) local details of the global map generated by our method. It is clear that our method effectively detect and filter moving objects, resulting in more consistent results.}
	\label{figure4}
\end{figure*}

\begin{table}[t]
	\caption{Map Quantitative Evaluation on KITTI Dataset}
	\centering
	\scalebox{1.0}{
		\begin{tabular}{|c|c|c|c|c|}
			\hline
			\rule{-1pt}{8pt}
			Seq. & Method & PR [$\%$] & RR [$\%$] & F1-Score \\
			
			\hline
			\rule{-1pt}{8pt}
			\multirow{5}{*}{KITTI00}
			& LIO-SEGMOT & 93.282 & 92.510 & 0.929 \\
			\cline{2-5}
			\rule{-1pt}{8pt}
			& LIMOT & 92.354 & 93.997 & 0.932 \\
			\cline{2-5}
			\rule{-1pt}{8pt}
			& RF-A-LOAM & 83.912 & 94.425 & 0.888 \\
			\cline{2-5}
			\rule{-1pt}{8pt}
			& Ours & \textbf{96.461} & \textbf{94.721} & \textbf{0.956} \\
			
			\hline
			\rule{-1pt}{8pt}
			\multirow{5}{*}{KITTI05}
			& LIO-SEGMOT & 91.485 & 86.134 & 0.887 \\
			\cline{2-5}
			\rule{-1pt}{8pt}
			& LIMOT & 90.950 & \textbf{90.452} & 0.907 \\
			\cline{2-5}
			\rule{-1pt}{8pt}
			& RF-A-LOAM & 87.931 & 85.333 & 0.866 \\
			\cline{2-5}
			\rule{-1pt}{8pt}
			& Ours & \textbf{92.450} & 89.315 & \textbf{0.909} \\
			
			\hline
			\rule{-1pt}{8pt}
			\multirow{5}{*}{KITTI07}
			& LIO-SEGMOT & 89.342 & 92.751 & \textbf{0.910} \\
			\cline{2-5}
			\rule{-1pt}{8pt}
			& LIMOT & 90.314 & 91.007 & 0.907 \\
			\cline{2-5}
			\rule{-1pt}{8pt}
			& RF-A-LOAM & 80.641 & \textbf{94.271} & 0.869 \\
			\cline{2-5}
			\rule{-1pt}{8pt}
			& Ours & \textbf{91.698} & 91.854 & 0.908 \\
			
			\hline
			
		\end{tabular}
	}
	\label{table2}
\end{table}

\subsection{Mapping Results} \label{sec4b}
1) Qualitative: As shown in Fig.~\ref{figure4}, we compare maps built by the proposed method and other baselines in a dynamic urban environment. In the map generated by DLO \cite{dlo}, we clearly see the severe \textit{ghosttail} in the map caused by moving cars. Comparatively, our method adeptly identifies and filters moving objects while retaining valuable semi-static objects. Compared with LIO-SEGMOT \cite{lio-segmot}, LIMOT \cite{limot} and RF-A-LOAM, our method shows better dynamic removal performance. Ultimately, we obtain a high-precision LiDAR odometry and build a cleaner map.

2) Quantitative: The evaluations on map quality are presented in Table.~\ref{table2}. The PR and RR in Table.~\ref{table2} represent the preserved rate of the static points and the removed rate of dynamic points. From the statistic, the PR scores of our method are higher than other baselines, while the RR scores are suboptimal to LIMOT \cite{limot}. Balancing PR and RR, the F1-Score of our method is competitive.

The motivation of this work is to achieve a compact dynamic LO system, devoting to improve the holistic accuracy of odometry and mapping through integrating object detection, tracking, and bounding box posture consistency constraint. We admit that we do not focus on how to further improve the performance of object detection and tracking. From the extensive comparison experiments, the proposed dynamic LO system is competitive against the baselines in overall consideration of localization accuracy, mapping quality and efficiency.

\begin{table}[t]
	\caption{Ablation Results Using RMSE of RPE (m) and APE (m) on KITTI07 and UrbanLoco05}
	\centering
	\scalebox{0.85}{
		\begin{tabular}{|c|ccc|cc|cc|}
			\hline
			\rule{-1pt}{10pt}
			\multirow{2}{*}{\normalsize{Version}} & \multirow{2}{*}{Tracker} & \multirow{2}{1cm}{\centering{\footnotesize{Dynamic} \\ \footnotesize{Removal}}} & \multirow{2}{1.2cm}{\centering{\footnotesize{Consistency} \\ \footnotesize{Constraint}}} & \multicolumn{2}{c|}{KITTI07} & \multicolumn{2}{c|}{UrbanLoco05}\\
			\cline{5-8}
			\rule{-1pt}{10pt}
			&  &  &  & RPE & APE & RPE & APE \\
			\hline
			\rule{-1pt}{8pt}
			(a) & UKF & \usym{2717} & \usym{2717} & 1.65 & 1.56 & 3.85 & 1.71 \\
			
			(b) & UKF & \usym{2717} & \usym{2714} & 1.52 & 1.45 & 3.79 & 1.27 \\
			
			(c) & UKF & \usym{2714} & \usym{2717} & 1.33 & 1.53 & 3.76 & 1.48 \\
			
			(d) & EKF & \usym{2714} & \usym{2714} & 1.24 & 1.23 & 3.56 & 1.29 \\
			
			(e) & UKF & \usym{2714} & \usym{2714} & \textbf{1.10} & \textbf{0.92} & \textbf{3.42} & \textbf{1.23} \\
			\hline
		\end{tabular}
	}
	\label{table3}
\end{table}

\begin{table}[t]
	\caption{Ablation Results of Ground Plane Segmentation Using RMSE of RPE (m) and APE (m), and Time (ms).}
	\centering
	\scalebox{0.95}{
		\begin{tabular}{|c|ccc|ccc|}
			\hline
			\rule{-1pt}{10pt}
			\multirow{2}{2.5cm}{\centering Ground Segmentation Method}  & \multicolumn{3}{c|}{KITTI07} & \multicolumn{3}{c|}{UrbanLoco05}\\
			\cline{2-7}
			\rule{-1pt}{10pt}
			& RPE & APE & Time & RPE & APE & Time \\
			\hline
			\rule{-1pt}{8pt}
			RANSAC (0.4m) & \textbf{0.99} & 0.95 & 24.50 & 3.73 & 1.25 &  3.49\\
			
			RANSAC (0.5m) & 1.19 & 1.06 & 16.29 & 3.81 & 1.30 & 2.77\\
			
			Ours & 1.10 & \textbf{0.92} & \textbf{0.37} & \textbf{3.42} & \textbf{1.23} & \textbf{0.36} \\
			\hline
		\end{tabular}
	}
	\label{table4}
\end{table}

\begin{figure}[tbp]
	\centering
	\includegraphics[width=0.485\textwidth]{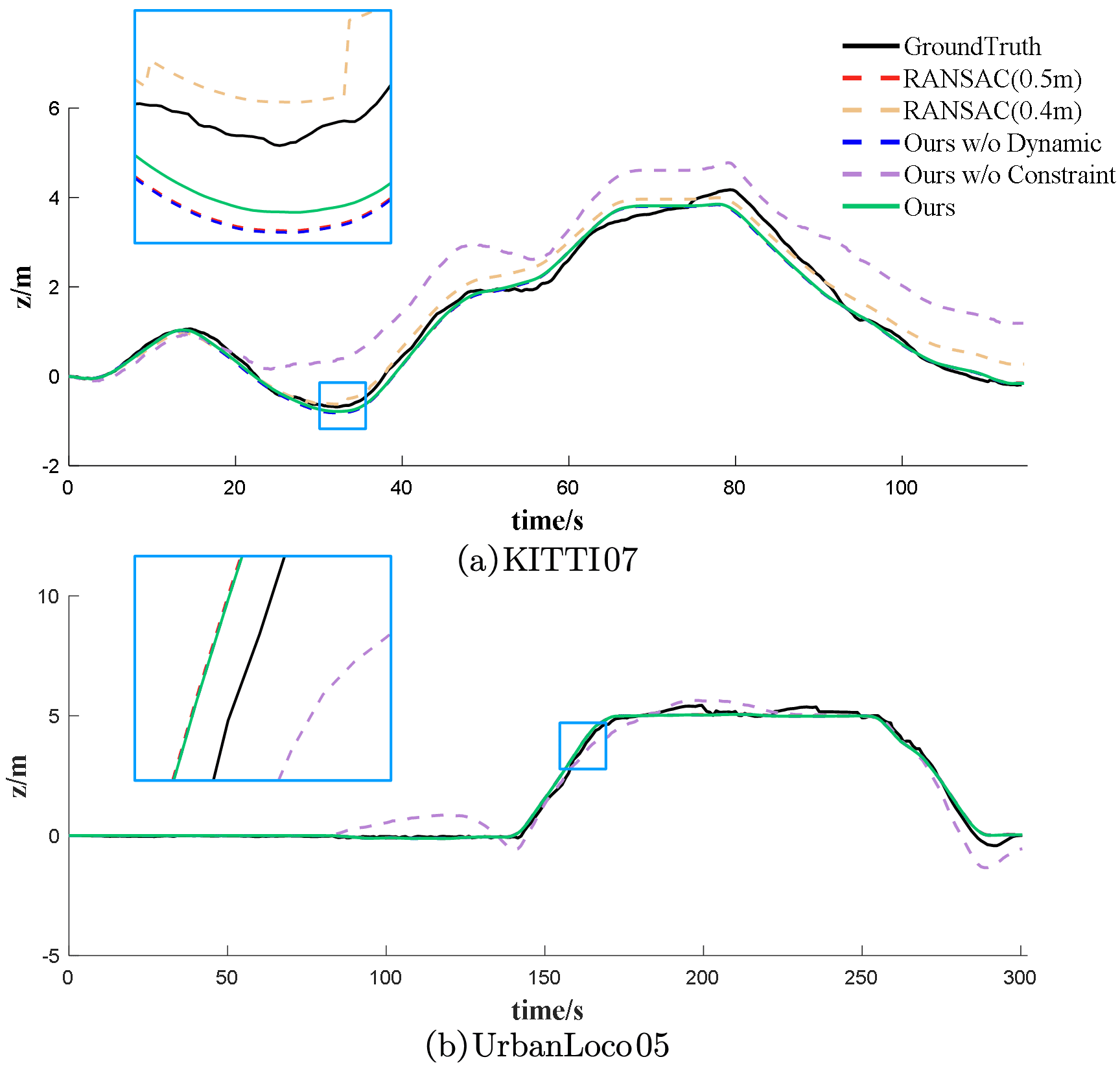}
	\caption{\textbf{z-axis drift of all methods on ablation study}. With imposing the bounding box consistency constraint, the z-axis drift is inhibited and the odometry accuracy is improved.}
	\label{figure5}
\end{figure}

\subsection{Ablation Study} \label{sec4c}
To investigate the effect of our system components, we conduct ablation studies on the KITTI07 and UrbanLoco05 sequences. We compare the accuracy of the following versions of the odometry: our method with no dynamic removal and no bounding box consistency constraint (a), our method with no dynamic removal (b), our method with no bounding box consistency constraint (c), our method that replaces UKF with EKF (d),  and our complete method (e). Additionally, according to the method mentioned in \cite{apt}, we use RANSAC for global segmentation to fit the ground and impose ground constraints. We also compare the effect of our posture consistency constraint with traditional ground segmentation method under different downsample sizes (0.4m, 0.5m).

1) Tracker: As shown in Table.~\ref{table3}, LO system with UKF-based object tracker outperforms EKF-based, benefiting from the more robust dynamic tracking. Theoretically, EKF addresses the nonlinear state estimation by using linear approximation, which may introduce tracking bias.

2) Dynamic Removal: As shown in Table~\ref{table3}, the results demonstrate that the outliers in the environment are effectively filtered by adding our dynamic removal step, resulting in a more accurate odometry estimation. This improvement is particularly obvious when dealing with the highly dynamic UrbanLoco dataset. 

3) Bounding Box Consistency Constraint: The results are shown in Fig.~\ref{figure5} and Table~\ref{table3}. The $z$-axis drift of the odometry is effectively inhibited by imposing the bounding box consistency constraint. Combined with dynamic removal and bounding box consistency constraint, our complete system achieves more accurate results. The translation accuracy of KITTI07 dataset and UrbanLoco05 dataset increases by 39.3\% and 30.4\%, respectively. 

As shown in Fig.~\ref{figure5} and Table~\ref{table4}, the proposed method with bounding box constraint achieves higher localization accuracy with less time cost. The reason for high efficiency is that the bounding box constraint liberates the step of ground segmentation, which is one of sub-contributions.

\begin{table}[t]
	\caption{Running Time (ms) Analyses for Processing one scan.}
	\centering
	\begin{tabular}{c|ccc|c}
		\toprule
		Method & Preprocessing & Tracker & Odometry & Total \\
		\midrule
		LIO-SEGMOT & 15.43 & 39.91 & 80.36 & 135.70 \\
		\midrule
		LIMOT & 13.74 & 9.83 & 65.11 & 88.68 \\
		\midrule
		Ours & 3.42 & 6.34 & 14.72 & 24.28 \\
		\bottomrule
	\end{tabular}
	\label{table5}
\end{table}

\subsection{Running Time Analysis} \label{sec4e}
As shown in Table~\ref{table5}, our approach achieves a time consumption twice faster than LIMOT \cite{limot} and five times faster than LIO-SEGMOT \cite{lio-segmot}. The proposed UKF-based 3D multi-target tracker takes less time to maintain the similar accuracy of dynamic removal compared with baselines. Besides high efficiency, our odometry module exhibits competitive localization accuracy.  In summary, our system can run over 20Hz, meeting real-time operation.

\section{Conclusion}

This work proposed TRLO, a LiDAR Odometry that is accurate and robust to dynamic environments for long-term traversals. One of our key innovations was to design a 3D multi-object tracker based on Unscented Kalman Filter and nearest neighbor method, which stably identifies and removes dynamic objects to provide the static scans for the two-stage iterative nearest point LiDAR odometry. In addition, an efficient hash-based keyframe management was proposed for the rapid submap construction. Finally, we reused 3D bounding boxes to further optimize the computed pose, obtaining the globally consistent pose and clean maps. Our benchmark tests conducted across various challenging dynamic environments confirmed the reliability of our approach. Through ablation study, we systematically explored the contribution of each component.

In the future, we are interested in integrating IMU into our system or adding loop closure to achieve more robust pose estimation.

\bibliographystyle{IEEEtran}
\bibliography{myRef.bib}

\vfill

\end{document}